\documentclass[letterpaper]{article} 
\usepackage[preprint]{aaai2027_jlu}  
\usepackage[hyphens]{url}  
\usepackage{graphicx} 
\usepackage{natbib}  
\usepackage{caption} 
\usepackage{eso-pic}
\usepackage{graphicx}
\usepackage{cite}
\usepackage{xcolor}
\usepackage{dsfont}
\usepackage{amsthm} 
\usepackage{booktabs} 
\usepackage{rotating}
\usepackage{multirow} 
\usepackage{amsmath}
\usepackage{amssymb}
\usepackage{mathtools}
\usepackage{amsthm}
\usepackage{multirow} 

\usepackage{algorithm}
\usepackage{algorithmic}

\usepackage{newfloat}
\usepackage{listings}
\DeclareCaptionStyle{ruled}{labelfont=normalfont,labelsep=colon,strut=off} 
\floatstyle{ruled}
\newfloat{listing}{tb}{lst}{}
\floatname{listing}{Listing}

\usepackage{booktabs}

\usepackage{tikz}
\usepackage{graphicx}

\makeatletter\def\@listi{\leftmargin\leftmargini \topsep .5em \parsep .5em \itemsep .5em}
\def\@listii{\leftmargin\leftmarginii \labelwidth\leftmarginii \advance\labelwidth-\labelsep \topsep .4em \parsep .4em \itemsep .4em}
\def\@listiii{\leftmargin\leftmarginiii \labelwidth\leftmarginiii \advance\labelwidth-\labelsep \topsep .4em \parsep .4em \itemsep .4em}\makeatother

\newcounter{checksubsection}
\newcounter{checkitem}[checksubsection]

\title{Beyond Flat Policies: Hierarchical Post-Training for Embodied Agents in Robotic Manipulation}
\author{
    He Kong \equalcontrib \textsuperscript{\rm 1}, Zengjue Chen \equalcontrib \textsuperscript{\rm 1}, Qi Wang \corresponding \textsuperscript{\rm 1}, Qianli Xing \textsuperscript{\rm 2}, Runliang Niu \textsuperscript{\rm 1}, Peidong Liu \textsuperscript{\rm 3}, Jiawei Li \textsuperscript{\rm 3}, Shiqi Wang \textsuperscript{\rm 1}, Yi Chang \textsuperscript{\rm 1}
\\
}
\affiliations{
    \textsuperscript{\rm 1}School of Artificial Intelligence, Jilin University\quad
    \textsuperscript{\rm 2}College of computer science and technology, Jilin University \\
    \textsuperscript{\rm 3}
    Joy Future Academy, JD \\
    konghe19@mails.jlu.edu.cn, \quad zengjue24@mails.jlu.edu.cn,\quad qiwang@jlu.edu.cn

}

\begin{document}
\maketitle

\begin{abstract}
Vision-language-action (VLA) models have demonstrated remarkable capabilities in robotic manipulation by leveraging pretrained vision-language models. However, existing post-training methods predominantly optimize VLA models as flat policies, making it difficult to explicitly model task progression and perform robust long-horizon manipulation. Although hierarchical approaches introduce task decomposition, they mainly rely on supervised learning from offline demonstrations and cannot effectively improve execution through online interaction. To address this limitation, we propose Hierarchical Robotic Control (HiRoC), a hierarchical post-training framework that decouples high-level task planning from low-level action execution. The planner decomposes complex tasks into executable subgoals to provide explicit semantic guidance, while the executor continuously improves subgoal-conditioned action generation through reinforcement learning. To enable effective collaboration between the two modules, we further align the executor with planner-generated subgoals before reinforcement learning, mitigating the distribution misalignment between planning and execution. Extensive experiments across diverse robotic manipulation benchmarks demonstrate that HiRoC consistently outperforms strong baselines. Comprehensive analyses further validate the effectiveness of hierarchical post-training and the contribution of each key component.
\end{abstract}

\section{Introduction}

Recently, vision-language-action (VLA) models have emerged as a promising paradigm for general-purpose robotic manipulation by integrating visual perception, language understanding, and action generation into a unified framework \cite{pmlrv229zitkovich23a,kim2024openvla,zhong2025survey,11495231}. 
These models have demonstrated strong capabilities across a wide range of manipulation tasks and robotic embodiments \cite{hu2023robofm,shukor2025smolvla}. Despite their impressive proficiency in short-horizon manipulation and primitive skills, real-world robotic tasks are often inherently long-horizon and multi-stage, requiring agents not only to generate accurate low-level actions but also to reason about \emph{what should be accomplished at each stage} \cite{belkhale2024rt,gao2025vla,long2026scaling}. Successfully completing such tasks therefore requires understanding task progression and decomposing a high-level instruction into a sequence of meaningful intermediate subgoals.

\begin{figure}[t]
\centering
\includegraphics[width=\linewidth]{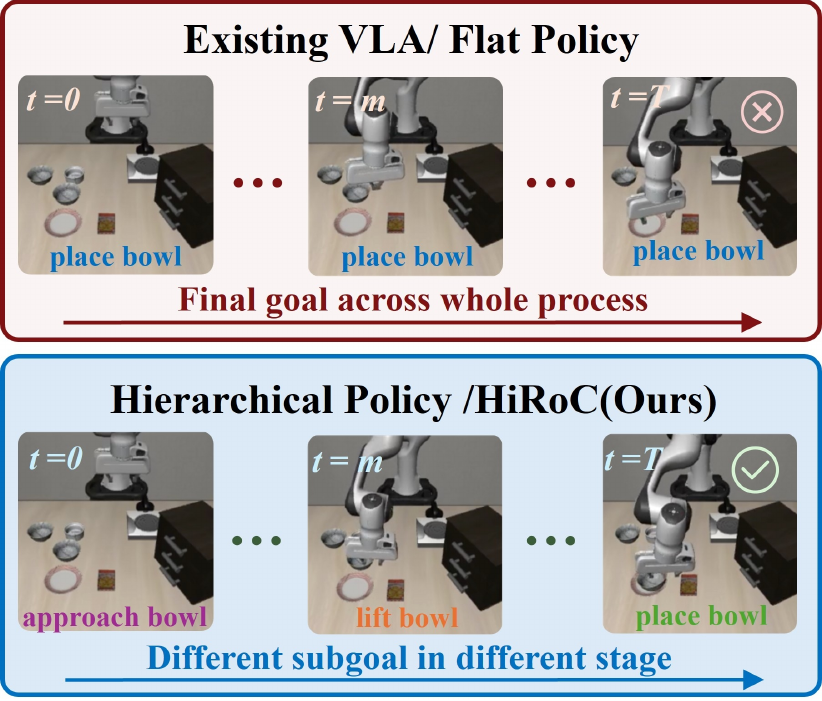}
\caption{\textbf{Motivation of HiRoC.} Existing flat VLA policies generate actions conditioned on the same global task instruction throughout execution.  HiRoC instead decomposes the global task into sequential subgoal, enabling more effective long-horizon decision-making.}
\label{motivation}
\end{figure}

However, existing VLA post-training methods still predominantly adopt a flat policy formulation, directly mapping visual observations and a global task instruction to low-level actions \cite{guo2025improving,lu2025vla,li2025simplevla,cao2026z}. As illustrated in Figure~\ref{motivation}, the same global instruction is typically maintained throughout execution, while the underlying stage-wise task structure remains implicit. 
In practice, complex tasks, especially long-horizon ones, consist of multiple stages with distinct semantic objectives. Consequently, conditioning only on the global instruction makes it difficult for the policy to identify its current stage and the next intermediate objective.
Moreover, minor execution deviations can accumulate over time without explicit subgoals serving as semantic anchors for subsequent decisions, eventually leading to degraded decision-making or task failure. Ultimately, flat post-training fails to endow VLA models with the hierarchical reasoning and self-correction capabilities essential for robust multi-stage execution.

A natural solution is to introduce an explicit planner that decomposes global instructions into intermediate subgoals, while using the VLA policy as a subgoal-conditioned executor. Although prior studies have explored hierarchical reasoning and structured planning, effectively post-training such a planner--executor hierarchy remains underexplored, as existing hierarchical methods largely rely on supervised learning while recent online RL methods still optimize flat VLA policies. This setting introduces three key challenges: the planner must generate correct and executable subgoals with appropriate transitions; the executor must overcome the distribution misalignment between global instructions and fine-grained subgoals, which otherwise causes a severe subgoal-conditioned execution cold start; and policy optimization must exploit intermediate subgoal progress rather than relying solely on trajectory-level outcomes. 

To address these challenges, we propose \textbf{Hi}erarchical \textbf{Ro}botics \textbf{C}ontrol (\textbf{HiRoC}), a hierarchical post-training framework that integrates task decomposition, planner--executor alignment, and online reinforcement learning for VLA models. HiRoC consists of a high-level planner that decomposes global task instructions into executable subgoals and a low-level VLA executor that generates actions conditioned on the current subgoal. Specifically, we first construct high-quality subgoal supervision data and train planner on these data by supervised fine tuning (SFT). To brige the misalignment between planner and executor, we perform subgoal-conditioned SFT on reorganized data to equip the executor with basic subgoal-following capabilities and mitigate the issue of cold start. 
Finally, we optimize the executor through online reinforcement learning under planner-generated subgoals and introduce a hierarchical Group Relative Policy Optimization (GRPO) objective  that incorporates subgoal-level progress into policy optimization, providing more structured learning signals for long-horizon execution. 
The contributions of our paper are summarized as follows:





$\bullet$ We propose HiRoC, a hierarchical post-training framework for VLA models that explicitly separates high-level task planning from low-level action execution. 
HiRoC introduces planner-guided subgoal decomposition to explicitly model task progression, enabling more structured reasoning and robust long-horizon robotic manipulation.

$\bullet$ We develop a collaborative training strategy for hierarchical VLA models. Specifically, we align planner-generated subgoals with the executor through subgoal-conditioned supervised fine-tuning to mitigate planner–executor distribution misalignment, and further introduce a hierarchical GRPO objective that jointly exploits task-level and subgoal-level learning signals for effective reinforcement learning.


$\bullet$ Extensive experiments demonstrate consistent performance improvements across diverse robotic manipulation benchmarks, achieving an average improvement of 10.06\%. Further analyses provide comprehensive insights into the effectiveness of hierarchical control and the contribution of each key component.

\section{Related Works}
\textbf{VLAs for Robotic Control.} Ever since that LLMs/VLMs achieved great process in multimodal understanding, researchers has investigated to integrate their backbones with action modules  for robotic manipulation tasks \cite{li2025simplevla}. R2-T \cite{pmlrv229zitkovich23a} firstly  proposes to co-fine-tune state-of-the-art VLMs on both robotic trajectory data and
Internet-scale vision-language tasks. 
CotVLA \cite{Zhao2025CoTVLAVC} further   incorporates explicit visual CoT reasoning into VLAs by predicting future image frames autoregressively as visual goals before generating a short action sequence to achieve these goals. In addition, 
OpenVLA \cite{kim2024openvla} utilizes parameter-efficient fine-tuning techniques to train VLAs on a diverse collection of 970k real-world robot demonstrations. 
VP-VLA \cite{wang2026vp}  decouples high-level reasoning and lowlevel execution via a structured visual prompting interface. 
VISTA \cite{long2026scaling} unifies world modeling and visual goal–conditioned
policy learning for long-horizon robotic manipulation by synthesizing textual subtasks and visual subgoals.  Some works like RT-H \cite{belkhale2024rt} and VLA-OS \cite{gao2025vla} investigate the idea of hierarchical task planning and action generation. 
Moreover, to overcome the data acquisition bottleneck, DexMimicGen \cite{11127809} automatically synthesizes trajectories from a
small number of human demonstrations for bimanual and
dexterous robot manipulation, then imitation learning is utilized to train VLA for robotic manipulation.
Although above works usually achieve satisfied performance by  conducting  imitation learning and SFT on pre-collected trajectories, high-quality data is too expensive to obtain good generalization on unseen scenarios.

\textbf{Reinforcement Learning for Robotic VLA Control.}
Reinforcement learning (RL) has played a fundamental role in embodied AI and is increasingly adopted to improve VLAs through interaction \cite{ma2024survey}.
GRAPE \cite{zhang2024grape} improves trajectory-level generalization by applying direct preference optimization to rewards implicitly derived from successful and failed trials.
RLinf \cite{yu2025rlinf} provides a flexible RL infrastructure for efficiently executing programmable training workflows.
Z-1 \cite{cao2026z} performs scene-specific SFT on RoboCasa demonstrations followed by task-wise GRPO post-training.
TGRPO \cite{chen2025tgrpo} integrates trajectory- and step-level information through group-based dual-level advantage estimation.
SimpleVLA-RL \cite{li2025simplevla} enables end-to-end online rule-based RL through VLA-specific trajectory sampling and loss computation.
iRe-VLA \cite{guo2025improving} combines SFT and RL fine-tuning to improve training stability and reduce computational overhead, while ActionX \cite{actionx} trains action experts via RL with a frozen pretrained VLM backbone.
However, existing RL-based methods often lack explicit task decomposition for long-horizon decision-making or fail to coordinate planning and execution.
We therefore propose a hierarchical manipulation framework that combines SFT and online RL, where a planner dynamically decomposes the task into subgoals and guides the executor toward accurate actions.

\begin{figure*}[h]
    \centering    \includegraphics[width=0.85\linewidth]{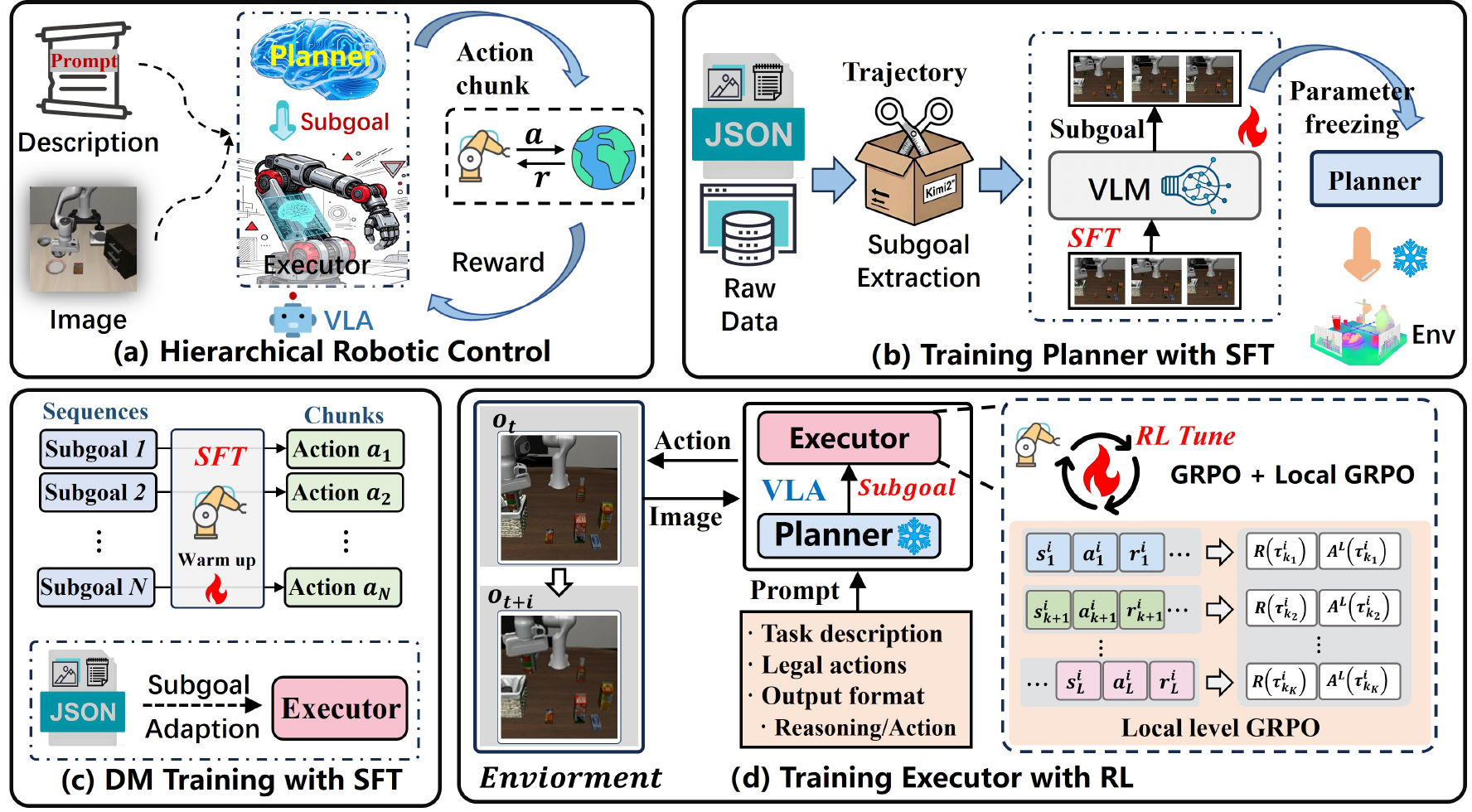}
    \caption{The framework of HiRoC. (a) shows the environmental interaction via our proposed HiRoC. The training process of HiRoC is divided into three parts: (b) train planner by SFT on pre-collected trajectories; (c) pretrain executor with SFT on prepared data to overcome the issue of distribution misalignment (DM); and (d) train executor by RL tuning when interacting with environment.}
    \label{framework}
\end{figure*}
\section{Preliminaries}
A VLA agent that interacts with an environment to accomplish multi-step tasks based on a task description $x \sim p(x)$ is considered in this work. The task of robotic manipulation by reinforcement learning can be modeled as a Markov decision process (MDP), which is defined by a tuple $\langle s, a, P, r, s', \gamma \rangle$. At each time step $t$, the agent observes a state $s_t \in S$, where we treat the combination of the vision ($\mathcal{O}$) and language inputs ($\mathcal{V}^m$) to VLAs as the state space: $S=\mathcal{O} \times \mathcal{V}^m$. $\mathcal{O}$ is the space of all RGB images, and $\mathcal{V}^m$ denotes the discrete and finite vocabulary (token) space, where $m$ is the maximum token length. The agent selects an action $a_t \in \mathcal{A}$ according to a policy $\pi_\theta(a_t \mid s_t, x)$ parameterized by $\theta$. The environment then returns a reward $r_t \in \mathds{R}$. In this work, a sparse terminal reward of 1 is assigned for success and 0 for failure. A full episode consisting of a trajectory $\boldsymbol{\tau}=\{(s_1,a_1,r_1),\cdots,(s_T,a_T,r_T)\}$ is saved in the replay buffer $D$. The objective of $\pi_\theta$ is to maximize the expected return with discount factor $\gamma$: $R=\sum_{t=1}^{T}\gamma^{t}r_t$. Specifically, HiRoC takes two steps to select an appropriate action: the planner generates the subgoal $l_t$ based on the current observation and the task description. The executor $\pi_\theta(a_t \mid o_t, l_t)$ then selects an action to interact with the environment.


\section{Methodology}
We propose a hierarchical decomposition framework for robotic manipulation control as shown in Figure \ref{framework}, named HiRoC. In HiRoC, the planner and executor are designed as two complementary modules, where the planner, trained via SFT, is responsible for generating subgoals according to the current situation. The generated subgoals are then utilized to guide the executor in making decisions, where SFT and RL tuning are employed to train the executor. 

\subsection{Training Planner by SFT}
Existing backbones for embodied AI fail to decompose tasks into subtasks and dynamically generate corresponding subgoals. To overcome this issue, we fine-tune our planner via SFT. However, the corresponding data are insufficient for training the planner. Thus, we clean and reorganize existing datasets for planner training. 

\textbf{Data preparation.} We reuse the multimodal data from VLA-OS \cite{gao2025vla}, including text and images, and organize them into a LlamaFactory-compatible format \cite{zheng2024llamafactory}. To address noisy and inconsistent annotations in the original data, we unify the label space, remove ambiguous or duplicate entries, and reannotate conflicting samples according to the planner's objective of predicting the immediate next subgoal. The resulting dataset, denoted as $\mathcal{D}_{\mathrm{plan}}$, is directly used for supervised fine-tuning.

\textbf{Training Planner.} 
Based on the prepared trajectories, the planner is optimized via supervised fine-tuning (SFT) to generate the corresponding subgoal conditioned on the current observation. Specifically, given the reconstructed dataset $\mathcal{D}_{\text{plan}}$, the SFT objective is formulated as follows:
\begin{equation}
\mathcal{L}^{\text{SFT}}_{\text{plan}}(\mathbf{w}_p)
=
-
\mathbb{E}_{(s_t,l_t)\sim\mathcal{D}_{\text{plan}}}
\left[
\sum_{k=1}^{|l_t|}
\log
P\!\left(
l_t^{(k)}
\,\middle|\,
s_t,
l_t^{(<k)};
\mathbf{w}_p
\right)
\right],
\end{equation}
where $s_t$ denotes the current multimodal observation consisting of visual and language inputs, and $l_t=\{l_t^{(1)},\ldots,l_t^{(|l_t|)}\}$ represents the target subgoal sequence. $l_t^{(<k)}$ denotes the previously generated tokens, and $\mathbf{w}_p$ denotes the planner parameters. By minimizing the autoregressive cross-entropy loss, the planner learns to generate the next subgoal conditioned on the current context, thereby improving its instruction-following and task decomposition capability. The planner is then frozen while optimizing the low-level executor. 


\subsection{Training Executor by SFT and RL Tuning}
With the planner capable of producing executable subgoals, the next challenge is enabling the executor to effectively interpret and follow these high-level instructions. However, a distribution mismatch arises because the executor was originally pre-trained on global goals rather than explicit subgoals, leading to the distribution misalignment problem before further RL tuning.

\textbf{Distribution Misalignment.}
Although the planner provides hierarchical subgoals, the pretrained executor is originally optimized to solve the complete task conditioned on the final goals rather than intermediate subgoals. Consequently, directly coupling the planner with the executor introduces a distribution misalignment, leading to a severe cold start problem before subsequent reinforcement learning. 
To alleviate this issue, we simultaneously reorganize the collected trajectories into an executor dataset $\mathcal{D}_{\text{exe}}$ during data preparation. Specifically, each trajectory is decomposed into multiple subgoal-level training samples $(s_i, l_i, \mathbf{a}_i)$, where $s_i$ denotes the multimodal observation corresponding to the current subgoal, $l_i$ is the associated subgoal, and $\mathbf{a}_i=\langle a_1,\cdots,a_m\rangle$ denotes the action chunk for accomplishing $l_i$. The executor is then initialized by supervised fine-tuning as follows:
\begin{equation}
\begin{aligned}
\mathcal{L}_{\mathrm{exe}}^{\mathrm{SFT}}(\mathbf{w}_e)
&=
-
\mathbb{E}_{(s_i,l_i,\mathbf{a}_i)\sim\mathcal{D}_{\mathrm{exe}}}\\
& \quad
\left[
\frac{1}{HD}
\sum_{t=1}^{H}
\sum_{d=1}^{D}
\log
P_{\pi}
\left(
\cdot \mid s_i,l_i
\right)
\right],
\end{aligned}
\end{equation}
where $\mathbf{w}_e$ denotes the parameters of the executor $\pi$, and $s_i$ is the multimodal observation. By minimizing the above objective, the executor learns to predict the  action sequence conditioned on the current observation and subgoal instead of the original task instruction, thereby effectively mitigating the cold-start issue and providing a better initialization for subsequent reinforcement learning.

\textbf{RL Tuning.}  
Once the executor has acquired basic subgoal-following capability, we further train it through RL tuning via online interaction with the environment guided by the high-level planner. Formally, the executor's policy can be expressed as follows: 
\begin{equation}
    \mathbf{a}_{i,t:t+H-1}
    \sim
    \pi_{\theta}
    \left(
        \cdot \mid o_{i,t}, l_{i,t}
    \right),
\end{equation}
where $\mathbf{a}_{i,t:t+H-1}$ denotes the predicted action chunk with horizon $H$. This hierarchical formulation decomposes a long-horizon manipulation task into a sequence of intermediate objectives that are easier for the low-level policy to execute. The online trajectories are collected into the on-policy rollout buffer $D$. 

During RL tuning, the high-level planner is treated as a frozen semantic planning module and is not updated through backpropagation. This design preserves the task decomposition capability of the pretrained VLM while focusing the optimization process on improving action generation under subgoal-conditioned observations.
To effectively optimize the subgoal-conditioned policy, we then adopt GRPO \cite{guo2025deepseek} as the underlying RL optimizer. For the same task instance or initial state, the current policy samples a group of $N$ trajectories. The cumulative reward of trajectory $i$ is defined as $R_i=\sum_{t=1}^{T_i}r_{i,t}$ over the trajectory length $T_i$. The trajectory rewards are then standardized within the group to obtain the task-level advantage, denoted by global GRPO :
\begin{equation}
    A_i^{\mathrm{task}}
    =
    \frac{R_i-\mu_g}{\sigma_g+\epsilon},
\end{equation}
where
\begin{equation}
    \mu_g
    =
    \frac{1}{N}\sum_{j=1}^{N}R_j,
    \qquad
    \sigma_g
    =
    \sqrt{
        \frac{1}{N}
        \sum_{j=1}^{N}
        \left(R_j-\mu_g\right)^2
    },
\end{equation}
and $\epsilon$ is a small constant introduced for numerical stability.

Since task-level rewards alone cannot fully capture the progress of intermediate subgoals, the subgoal structure generated by the planner and recorded during rollout provides an additional learning signal.
Thus, we propose a local-level GRPO based on these subtasks to augment the global GRPO. Let $S_i$ denote the aggregated subtask score of trajectory $i$, computed from the intermediate subgoal segments contained in that trajectory. The corresponding subtask-level advantage is obtained through group-wise standardization:
\begin{equation}
    A_i^{\mathrm{sub}}
    =
    \frac{S_i-\mu_s}{\sigma_s+\epsilon},
\end{equation}
where
\begin{equation}
    \mu_s
    =
    \frac{1}{N}\sum_{j=1}^{N}S_j,
    \qquad
    \sigma_s
    =
    \sqrt{
        \frac{1}{N}
        \sum_{j=1}^{N}
        \left(S_j-\mu_s\right)^2
    }.
\end{equation}

To jointly exploit both sources of supervision, 
the final advantage used for optimizing executor is defined as follows:
\begin{equation}
    \widetilde{A}_i
    =
    w_t A_i^{\mathrm{task}}
    +
    w_s A_i^{\mathrm{sub}},
\end{equation}
where $w_t$ and $w_s$ control the contributions of the global task-completion signal and the intermediate subgoal-level signal, respectively. 
Since different tasks contain different numbers of intermediate subgoals, the contributions of the two advantages should be adaptively balanced as follows:
\begin{equation}
    w_t = \frac{K+L}{L},
    \qquad
    w_s = \frac{K}{K+L},
    \label{weighting}
\end{equation}
where
\begin{equation}
    K = \frac{1}{N}\sum_{i=1}^{N}K_i
\end{equation}
is the average number of recorded subgoals per trajectory, and $L$ denotes the length of collected trajectories. By incorporating multi-level trajectory importance, this weighting mechanism yields more effective policy learning.

\begin{table*}[htbp]
  \centering
  \small
  \begin{tabular}{c|cccccccc|cc}   
    \toprule
    \multirow{2}{*}{Method} &
    \multicolumn{2}{c}{LIBERO-Spatial} &
    \multicolumn{2}{c}{LIBERO-Object} &
    \multicolumn{2}{c}{LIBERO-Goal} &
    \multicolumn{2}{c}{LIBERO-Long} &
    \multicolumn{2}{c}{Average} \\
    \cmidrule(lr){2-3} \cmidrule(lr){4-5} \cmidrule(lr){6-7} \cmidrule(lr){8-9} \cmidrule(lr){10-11}
          & SR    & Rank  & SR    & Rank  & SR    & Rank  & SR    & Rank  & SR    & Rank \\
    \midrule
    OpenVLA\cite{kim2024openvla} & 84.7  & 1.0   & 88.4  & 2.0   & 53.7  & 1.0   & 79.2  & 3.0   & 76.5{\textcolor{black}{$\uparrow$17.0}}  & 1.8  \\
    OpenVLA*-Full & 91.6  & 9.0   & 95.3  & 8.0   & 90.6  & 10.0   & 86.5  & 7.0   & 91.0{\textcolor{black}{$\uparrow$2.5}}  & 8.5  \\
    SmolVLA\cite{shukor2025smolvla}& 93.0  & 10.0  & 94.0  & 7.0   & 77.0  & 6.0   & 91.0 & 10.0 & 88.8{\textcolor{black}{$\uparrow$4.7}}  & 8.3  \\
    ThinkAct\cite{huang2026thinkact} & 88.3  & 5.0   & 91.4  & 3.0   & 70.9  & 5.0   & 87.1  & 8.0  & 84.4{\textcolor{black}{$\uparrow$9.1}}  & 5.3  \\
    TGRPO\cite{chen2025tgrpo} & 90.4  & 8.0   & 92.2  & 6.0   & 81.0  & 8.0   & 59.2  & 1.0   & 80.7{\textcolor{black}{$\uparrow$12.8}}  & 5.8  \\
    MolmoAct\cite{lee2025molmoact} & 87.0  & 2.5   & 95.4  & 9.0   & 77.2  & 7.0   & 87.6  & 9.0   & 86.6{\textcolor{black}{$\uparrow$6.9}}  & 6.9  \\
    World-Env\cite{xiao2025world} & 87.6  & 4.0   & 86.6  & 1.0   & 57.8  & 3.0   & 86.4  & 6.0   & 79.6{\textcolor{black}{$\uparrow$13.9}}  & 3.5
    \\
    GRAPE\cite{zhang2024grape} & 88.5  & 6.0   & 92.1  & 5.0   & 57.2  & 2.0   & 83.1  & 5.0   & 80.2{\textcolor{black}{$\uparrow$13.3}}  & 4.5  \\
    VAL-OS-A-S\cite{gao2025vla} & 87.0  & 2.5   & \textbf{96.5} & \textbf{11.0} & \textbf{92.7}  &\textbf{10.0}  & 66.0  & 2.0   & 85.6{\textcolor{black}{$\uparrow$7.9}}  & 6.6  \\
    VLA-RL\cite{lu2025vla} & 90.2  & 7.0   & 91.8  & 4.0   & 59.8  & 4.0   & 82.2  & 4.0   & 81.0{\textcolor{black}{$\uparrow$12.5}}  & 4.8  \\
    \midrule
    HiRoC (Ours)  & \textbf{95.6} & \textbf{11.0} & 96.0  & 10.0  & 84.4 & 9.0 & \textbf{98.0}  & \textbf{11.0}   & \textbf{93.5} & \textbf{10.3} \\
    \bottomrule
  \end{tabular}%
  \caption{Performance comparison on LIBERO benchmark.  OpenVLA* incorporates action chunking and parallel decoding on the basis of
OpenVLA.  Success rate is utilized and $\uparrow$ indicates performance gain over the baseline. Numbers in bold denote the best performance.}
 \label{overallperformance}%
\end{table*}

Based on the resulting hierarchical advantage, we optimize the low-level executor's policy using a PPO-style clipped actor objective. Specifically, during training, the current policy recomputes the log probabilities of the action tokens sampled during rollout. Then, the probability ratio between the current and behavior policies is defined as follows:
\begin{equation}
\begin{aligned}
\rho_{i,t,j}(\theta)
&=
\exp \Big[
    \log \pi_{\theta}
    \big(
        a_{i,t,j}
        \mid o_{i,t}, l_{i,t}, a_{i,t,<j}
    \big)
    \\
&\qquad
    -
    \log \pi_{\theta_{\mathrm{old}}}
    \big(
        a_{i,t,j}
        \mid o_{i,t}, l_{i,t}, a_{i,t,<j}
    \big)
\Big].
\end{aligned}
\label{eq:policy_ratio}
\end{equation}
where $j$ indexes an action token in the predicted action chunk and $\theta_{\mathrm{old}}$ denotes the policy used to collect the rollout data. 
Finally, the loss of our executor is calculated as follows:
\begin{equation}
\begin{aligned}
\mathcal{L}_{\mathrm{actor}}(\theta)
&=
-\frac{1}{\sum_{i,t,j} m_{i,t,j}}
\sum_{i,t,j} m_{i,t,j}
\\
&\quad \times
\min \Big[
    \rho_{i,t,j}(\theta)\widetilde{A}_{i},
\\
&\qquad
    \operatorname{clip}
    \big(
        \rho_{i,t,j}(\theta),
        1-\epsilon_{\ell},
        1+\epsilon_{u}
    \big)
    \widetilde{A}_{i}
\Big].
\end{aligned}
\label{eq:actor_loss}
\end{equation}
where $\epsilon_{\ell}$ and $\epsilon_u$ are the lower and upper clipping thresholds, respectively. The binary mask $m_{i,t,j}\in\{0,1\}$ identifies valid action tokens, ensuring that only action tokens contribute to the optimization objective. The clipping operation limits the magnitude of each policy update and prevents the updated policy distribution from deviating excessively from the behavior policy.

\textbf{Remark.} The planner remains frozen during RL tuning for two reasons. Sparse trajectory rewards provide ambiguous credit assignment between planning and execution, making planner optimization unreliable. Moreover, updating the planner continuously shifts the subgoal distribution, breaking the planner-executor alignment established during SFT and exacerbating the distribution misalignment. Therefore, we optimize only the executor during RL fine-tuning.

\section{Experiments}
\textbf{Experimental Setup.}
To validate the effectiveness of HiRoC, we conduct experiments on LIBERO~\cite{liu2023libero} using its \emph{Goal}, \emph{Spatial}, \emph{Object}, and \emph{Long} suites. Each suite has 10 tasks evaluating distinct capability aspects. To evaluate generalization, we  conduct experiments on LIBERO-Plus~\cite{fei2025liberoplus}, which adds seven perturbations.

We adopt Qwen2.5-VL-3B as the planner and fine-tune it with LoRA on our prepared data. OpenVLA-OFT is employed as the executor. The executor is first fine-tuned via supervised fine-tuning (SFT) with LoRA on the reorganized subgoal-conditioned data, followed by reinforcement learning tuning. The group size of GRPO is set to 8. During evaluation, we perform 50 test episodes for each task and report the average success rate for each task and benchmark suite. All experiments are conducted on 8 NVIDIA H200 GPUs. We compare HiRoC with ten representative baselines. More details are provided in the Appendix.

\begin{figure*}
    \centering
    \includegraphics[width=\linewidth]{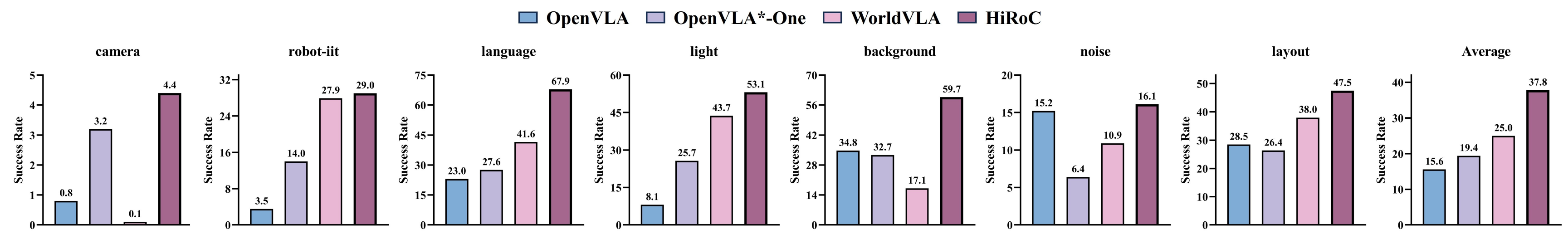}
    \caption{Performance of compared baselines in terms of zero-shot.}
    \label{zeroshot}
\end{figure*}
\subsection{Overall Performance}
Table \ref{overallperformance} shows the average success rates and ranks on LIBERO. The results demonstrate that HiRoC achieves state-of-the-art (SOTA) performance. Specifically, compared with RL-based methods such as VLA-RL, HiRoC adopts a hierarchical RL framework in which the planner provides more informative guidance to the executor, leading to superior performance. This observation is further supported by VAL-OS-A-S, which also incorporates explicit planning. World-model-based methods typically model environment dynamics or learn reward models to improve decision-making. However, HiRoC outperforms both World-Env and GRAPE because, although they better understand the environment, they cannot explicitly decompose complex tasks into sequential subgoals. TGRPO augments GRPO with step-level advantage estimation to better capture long-horizon task structures. In contrast, HiRoC augments GRPO with planner-generated subgoal trajectories, providing more meaningful intermediate supervision for policy optimization. Moreover, HiRoC achieves a 98\% success rate on the \emph{Long} suites, further demonstrating the effectiveness of hierarchical control for complex manipulation tasks.
This also explains why HiRoC obtains larger gains on Long and Spatial suites, where explicit subgoal decomposition is more beneficial, while Object and Goal suites rely more on visual grounding and direct goal understanding, respectively.

\subsection{Transfer Ability (Zero-shot Performance)}

To evaluate the generalization capability of HiRoC , we validate it on LIBERO-Plus, where OpenVLA, OpenVAL*-One, and WorldVLA are compared. Figure \ref{zeroshot} shows the success rate of each perturbation and average success rate. 
HiRoC achieves the best generalization across all perturbation types. Its substantial improvement over SFT-based models such as OpenVLA indicates that online interaction explores more diverse trajectories than a static full-shot dataset. Although WorldVLA improves generalization by modeling environment dynamics, it remains sensitive to environmental changes and requires additional data to adapt to new tasks. In contrast, HiRoC decomposes the goal into subgoals, allowing the executor to focus on each subgoal and thereby reducing its burden.




\subsection{Analysis of Distribution Misalignment}
\begin{figure}
    \centering
    \includegraphics[width=\linewidth]{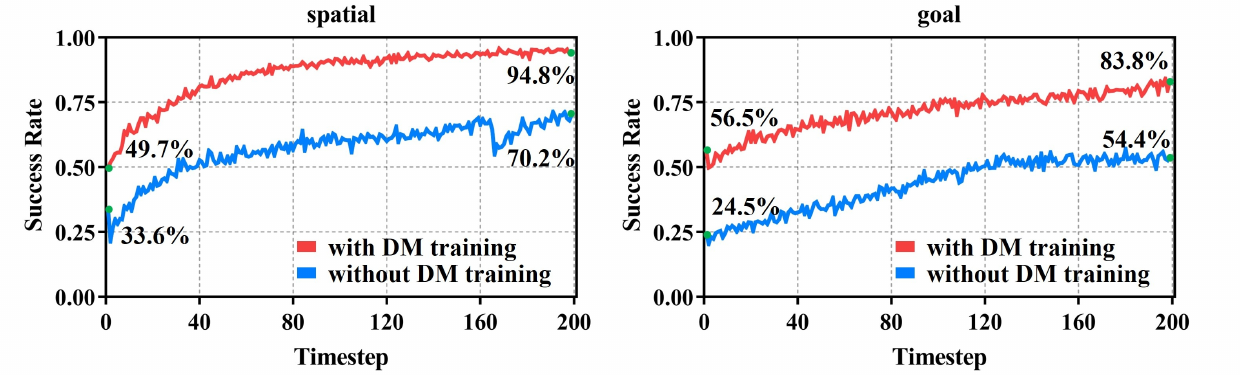}
    \caption{Learning curves of HiRoC with and without DM training in terms of rollout's success rate.}
    \label{coldstart}
\end{figure}

Before RL tuning, we observe that the executor cannot effectively execute the planner-generated subgoals due to the distribution misalignment (DM) between the planner and the executor. To alleviate this issue, we pretrain the executor on the prepared data via SFT.
To demonstrate this issue and verify the necessity of DM training, Figure \ref{coldstart} presents the learning curves of HiRoC with and without DM training on \emph{spatial} and \emph{goal}. We observe that, without DM training, the initial success rate is significantly lower. Although subsequent RL tuning further improves the policy, the final performance remains unsatisfactory. In contrast, DM training enables the executor to better understand planner-generated subgoals, resulting in superior performance.
\begin{figure}
    \centering
    \includegraphics[width=0.95\linewidth]{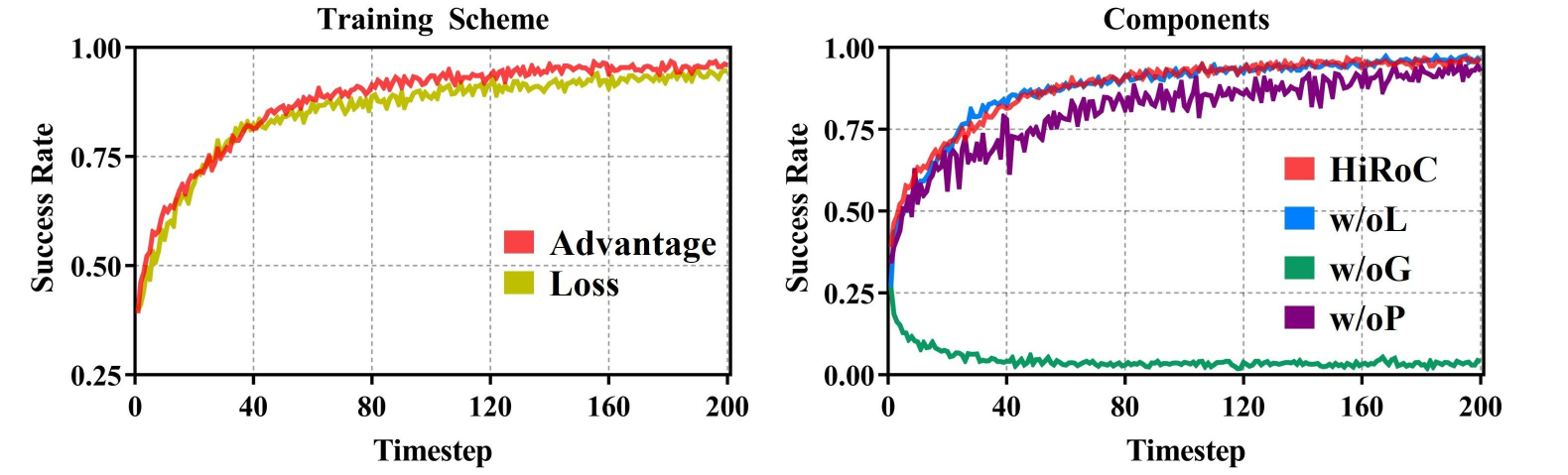}
    \caption{Learning curves on \emph{Object}.}
    \label{Fablation}
\end{figure}

\subsection{Analysis of Training Scheme}
HiRoC augments global GRPO with local GRPO over subtask trajectories. We compare two integration strategies: directly summing their losses or combining their advantage functions, as adopted in HiRoC. As shown on the left of Figure~\ref{Fablation}, advantage-based integration achieves better performance on \emph{Object}. This is likely because full and subtask trajectories induce different optimization objectives, making direct loss summation prone to suboptimal convergence. In contrast, advantage-level integration alleviates this inconsistency through the component-wise weighting in Eq.~(\ref{weighting}).
\begin{table}
  \centering
\small
 
    \begin{tabular}{c|c|cc|c}
    \toprule[1.15pt]
    L. & Method & Subgoal & Episode & Avg. \\
    \midrule
    \multicolumn{1}{c|}{\multirow{3}[2]{*}{\begin{sideways}Short\end{sideways}}} & HiRoC   & 94.92  & 91.55 & 93.24 \\
          & Robobrain2-3B & 0.04    & 0.00   & 0.02 \\
          & Robobrain2-7B & 0.29  & 0.00   & 0.15 \\
    \midrule
    \multicolumn{1}{c|}{\multirow{3}[2]{*}{\begin{sideways}
        Medium
    \end{sideways}}} & HiRoC   & 95.20 & 84.01 & 89.61 \\
          & Robobrain2-3B & 0.06    & 0.00   & 0.03 \\
          & Robobrain2-7B & 0.70    & 0.00   & 0.35 \\
    \midrule
    \multicolumn{1}{c|}{\multirow{3}[2]{*}{\begin{sideways}
        Long
    \end{sideways}}} & HiRoC   & 95.09  & 80.40 & 87.75 \\
          & Robobrain2-3B & \multicolumn{1}{c}{0.06}    & 0.00  & 0.03 \\
          & Robobrain2-7B & 0.98    & 0.00   & 0.49 \\
    \midrule
    \multicolumn{1}{c|}{\multirow{3}[2]{*}{\begin{sideways}
        Mix
    \end{sideways}}} & HiRoC   & 95.07  & 85.32 & 90.20 \\
          & Robobrain2-3B & 0.06    & 0.00   & 0.03 \\
          & Robobrain2-7B & 0.66  & 0.00   & 0.33 \\
    \bottomrule[1.15pt]
    \end{tabular}%

   \caption{Cross-validation results of HiRoC. Trajectories are divided into short, medium, and long parts according to their length. Results of mix denotes the mean of all trajectories. }
    \label{Planner_validatoin}%
\end{table}%


\begin{figure*}[ht] 
    \centering 
    \begin{minipage}{0.3\textwidth} 
        \centering
        \includegraphics[width=\textwidth]{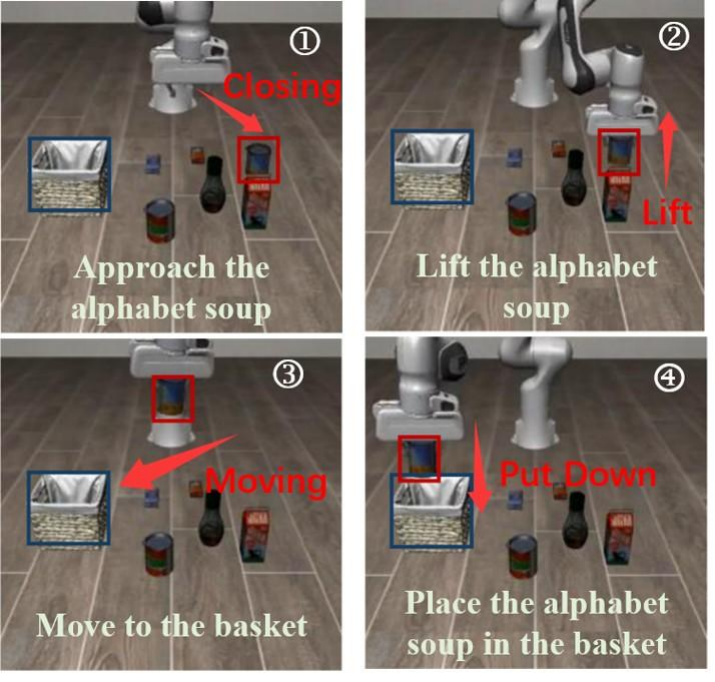} 
        \caption{The case study.}
        \label{case1}
    \end{minipage}\hfill 
    \begin{minipage}{0.7\textwidth} 
        \centering
        \includegraphics[width=\textwidth]{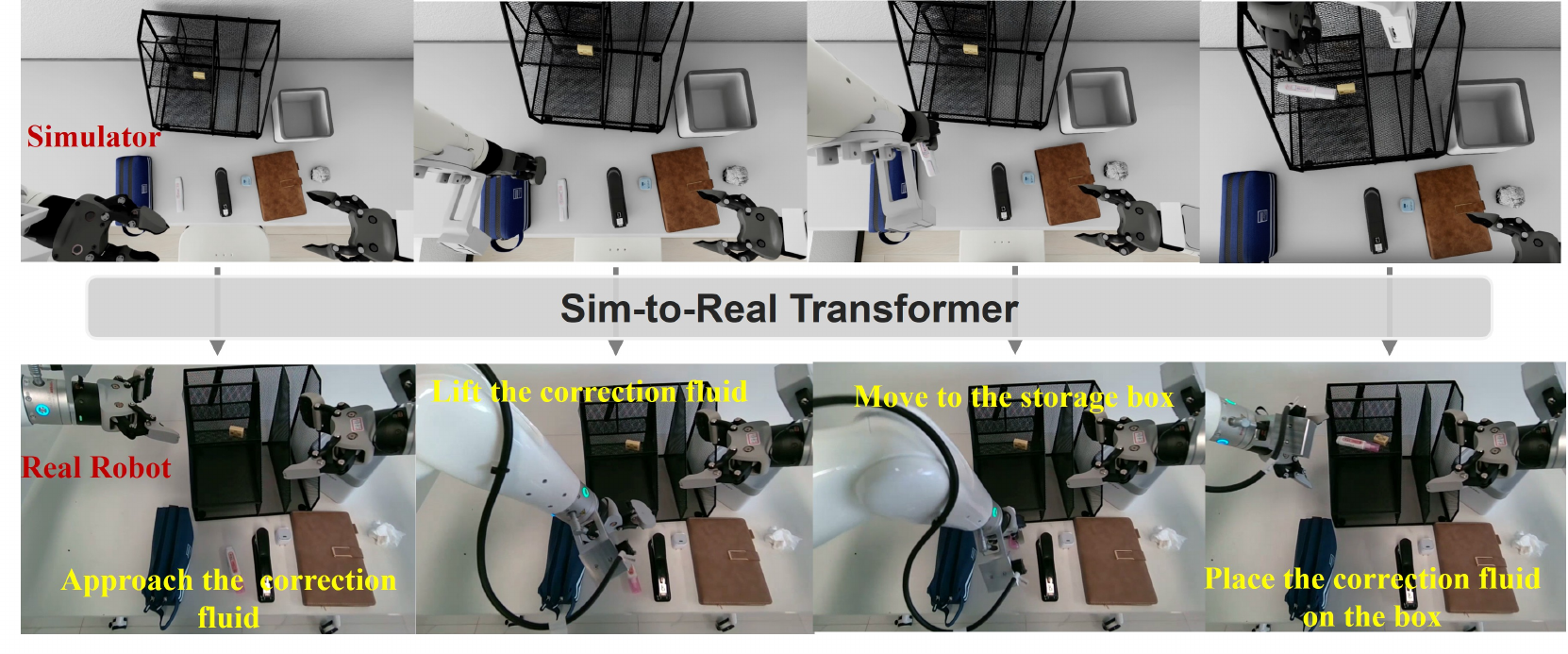} 
        \caption{The real world experiment: put correction fluid in box layer 2.}
        \label{simtoreal}
    \end{minipage}
\end{figure*}


\subsection{Analysis of Planner}
HiRoC trains a Qwen2.5-VL-3B  planner to decompose high-level tasks into subgoals. 
To validate its effectiveness and necessity, we perform 5-fold cross-validation and evaluate the planner using two metrics, namely text similarity and overall similarity, across episodes of different lengths. The results are reported in Table \ref{Planner_validatoin}, where 
RoboBrain2-3B and RoboBrain2-7B \cite{RoboBrain2.0TechnicalReport} are compared. They are designed for multi-step task planning.
The results in Table \ref{Planner_validatoin} show that HiRoC-Planner consistently achieves near-perfect accuracy across episodes of different lengths, whereas the state-of-the-art RoboBrain2 fails to generate accurate plans under the same settings. Since the planner provides hierarchical guidance for the downstream executor, its performance directly affects the overall control policy. Therefore, training the planner in advance is essential for reliable and accurate task execution.

\subsection{Case Study}


Figure \ref{case1} illustrates an example from \emph{Long}, where HiRoC is required to place the alphabet soup from the ground into the basket. The high-level planner decomposes the task into four subgoals. Initially, it instructs the executor to approach the alphabet soup. Once the robotic arm reaches the target, the planner updates the subgoal to lift the alphabet soup, and the executor adjusts its actions accordingly. After grasping the soup, the planner guides the executor to move toward the basket. Finally, the planner changes the subgoal to place the alphabet soup into the basket, and the executor successfully completes the task. By decomposing the complex task into several simple subtasks and providing accurate subgoal guidance, HiRoC effectively accomplishes the task, demonstrating the effectiveness and necessity of hierarchical control.

\subsection{Real World Experiments}

To validate the practical applicability of HiRoC, we conduct a sim-to-real experiment on a real robotic platform, as shown in Figure \ref{simtoreal}. During training, the executor is optimized  entirely in a high-fidelity simulator, where the workspace, object dimensions, and robot configuration are consistent with those of the real system.
The trained VLA policy is then directly deployed to the real robot without additional fine-tuning. Guided by the planner-generated subgoals, the executor successfully approaches, grasps, transports, and places the target object. The successful deployment demonstrates the effectiveness of HiRoC for real-world robotic manipulation and its promising sim-to-real transfer capability.


\subsection{Ablation Study}
In this subsection, ablation studies are conducted from two aspects: 1) we investigate the effectiveness of the planner (w/oPlan); and 2) we analyze the contributions of global GRPO (w/oG) and local GRPO (w/oL) to demonstrate their complementary roles. 
As shown in Table \ref{TAbaltionStudy}, removing the planner significantly degrades the final performance. Regarding GRPO, the original GRPO plays the primary role, while local GRPO further improves the performance of HiRoC. The right side of Figure \ref{Fablation} shows the learning curves of different variants of HiRoC, further supporting the above analysis. In addition, the planner also contributes to more stable learning. Without the planner, HiRoC becomes unstable and the learning curve exhibits severe fluctuations. Therefore, the proposed components play different yet complementary roles, enabling HiRoC to learn effective policies.
\begin{table}[tbp]
  \centering
    \begin{tabular}{c|cccc}
    \toprule
    Methods & w/oL & w/oG & w/oP & HiRoC \\
    \midrule
    success rate & 95.20\% & 4\%   & 92.60\% & 96\% \\
    \bottomrule
    \end{tabular}%
  \caption{Ablation study on \emph{Object}.
   \label{TAbaltionStudy}%
   }
\end{table}%


\section{Conclusion}

In this paper, we propose HiRoC, a hierarchical post-training framework with a high-level planner and a low-level executor. The planner, trained via SFT, decomposes complex tasks into simpler subtasks and generates subgoals to guide execution. To address the distribution misalignment between the two modules, we reorganize the subgoal-conditioned dataset and further optimize the executor through RL tuning. Experimental results and extensive analyses demonstrate the effectiveness of HiRoC and each of its components.  Although HiRoC demonstrates the value of hierarchical decomposition for robotic manipulation, future work will explore end-to-end training and more robust decomposition mechanisms for handling uncertainty in real-world physical systems.  
We hope HiRoC underscores the importance of combining hierarchical task decomposition with reinforcement learning-based policy optimization for complex robotic manipulation.


\bibliography{aaai2027}

\section{Appendix}

\subsection{Experimental Details}
\textbf{Simulation experiments.}
We conduct all experiments on the four LIBERO benchmark suites, including Spatial, Object, Goal, and the long-horizon LIBERO-10 suite. Following prior work, we first fine-tune the pretrained OpenVLA-OFT model on a small subset of demonstrations from each suite to obtain a suite-specific base executor. The high-level planner is initialized from Qwen2.5-VL-3B and fine-tuned via supervised fine-tuning (SFT) on our curated planner dataset to generate intermediate language subgoals. 

During HiRoC training, the low-level executor is initialized from the corresponding suite-specific OpenVLA-OFT checkpoint, while the planner remains frozen throughout RL tuning. The planner generates language subgoals conditioned on the current observation and task instruction, and replanning is performed every 20 policy calls. We adopt the GRPO with a group size of 64 vectorized environments per worker, 16 rollout epochs for each policy update, and 512 environment steps per trajectory. The executor predicts action chunks of length 8 and is trained using BF16 Fully Sharded Data Parallel (FSDP) with a learning rate of $2\times10^{-5}$. Action normalization statistics are computed separately for each LIBERO suite. We train the policy for 200 epochs on Spatial, Object, and Goal, and for 150 epochs on the long-horizon LIBERO-10 suite. During evaluation, each episode has a maximum horizon of 512 environment steps, and the average task success rate is reported. All experiments are conducted using eight deterministic environment seeds, with one seed assigned to each GPU worker.

\textbf{Real world experiments.} We conduct the real world experiments in JoySim using JoyRA-0.1 as the base VLA model. The task requires the robot to grasp a correction fluid bottle from the tabletop and place it into a box. The high-level planner is initialized from Qwen2.5-VL-3B and fine-tuned via supervised fine-tuning (SFT), while the low-level executor is initialized from JoyRA-0.1 and optimized using Flow-SDE through reinforcement learning in simulation. After training, the learned policy is directly deployed to the real robotic platform for zero-shot evaluation without any additional real-world fine-tuning.

\subsection{Prompt Details}
\begin{figure}[t]
    \centering
    \includegraphics[width=\linewidth]{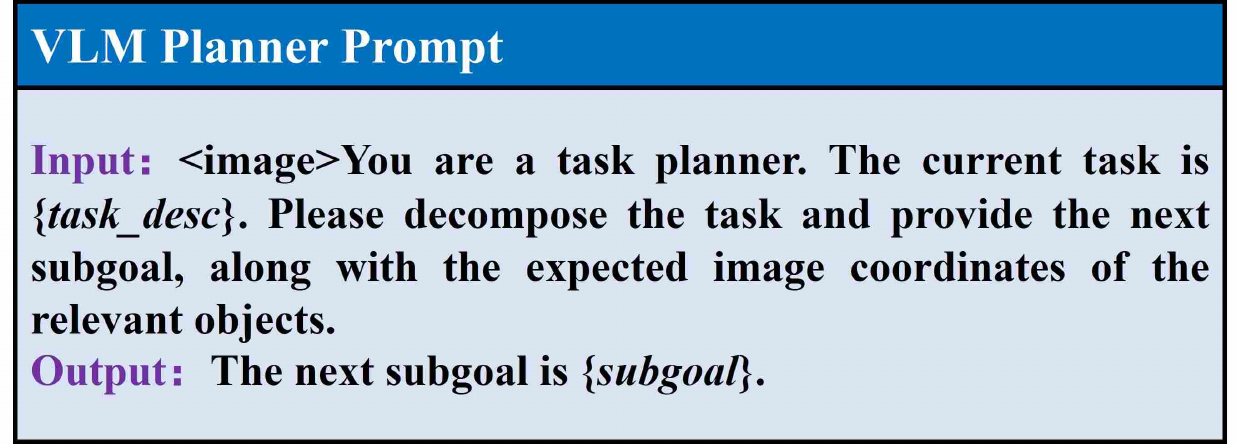}
    \caption{The prompt of high-level planner.}
    \label{p1}
\end{figure}

\begin{figure}[t]
    \centering
    \includegraphics[width=\linewidth]{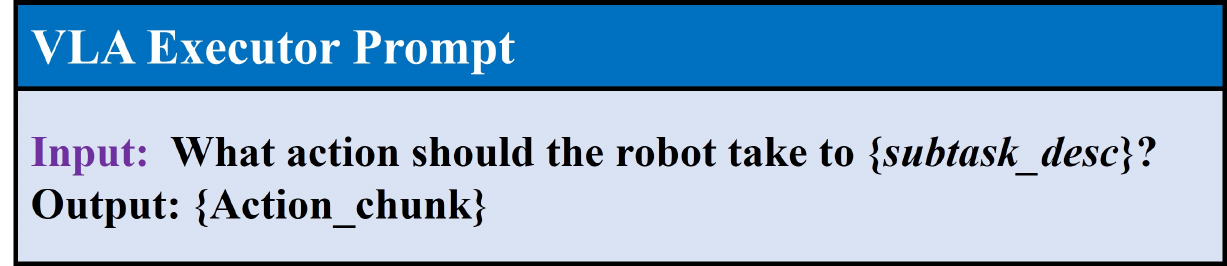}
    \caption{The prompt of low-level executor.}
    \label{p2}
\end{figure}

Figures \ref{p1} and \ref{p2} show the prompts for high-level planner and low-level executor.


\begin{figure*}[t]
    \centering
    \includegraphics[width=\linewidth]{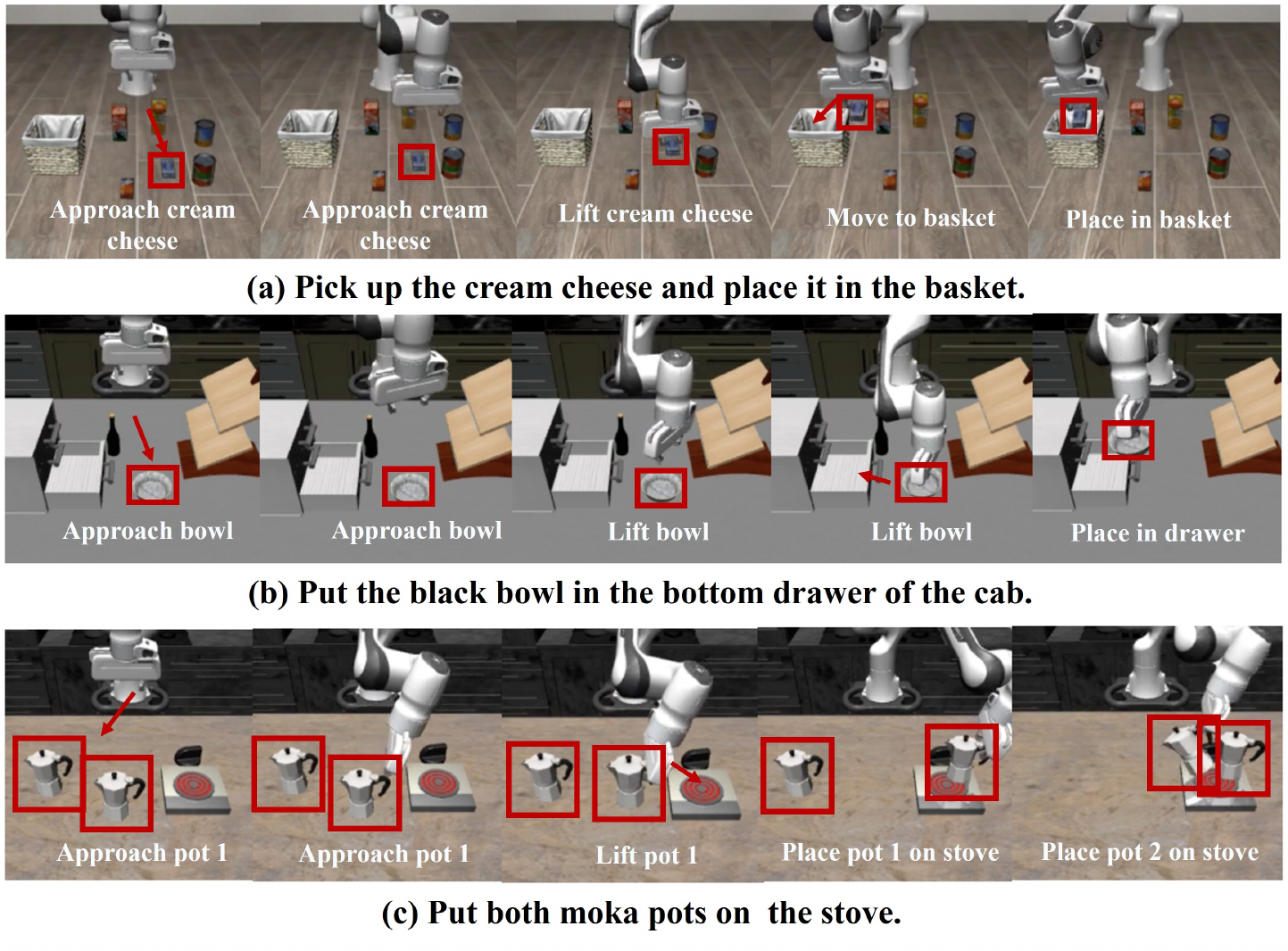}
    \caption{Case studies from LIBERO.}
    \label{case}
\end{figure*}
\subsection{Case Study}
Figure~\ref{case} presents several representative manipulation cases to
illustrate the effectiveness of HiRoC in long-horizon robotic tasks. 
Unlike flat VLA policies that continuously condition on the global instruction,
HiRoC explicitly decomposes complex tasks into sequential subgoals through the
high-level planner, allowing the executor to focus on the current manipulation
objective.

In Figure~\ref{case}(a), the robot is required to pick up the cream cheese
and place it into the basket. The planner first generates the subgoal
``approach cream cheese'', guiding the executor to locate and reach the target
object. After the object is approached, the planner updates the objective to
``lift cream cheese'', enabling the executor to perform accurate grasping
actions. Subsequently, the subgoal is changed to ``move to basket'' and
``place in basket'', respectively. By providing stage-specific semantic
guidance, HiRoC avoids the ambiguity of directly executing the complete task
instruction and successfully completes the long-horizon manipulation sequence.

Figure~\ref{case}(b) further demonstrates the ability of HiRoC to handle
tasks requiring sequential interactions with different spatial targets. For the
instruction of placing the bowl into the bottom drawer, the planner gradually
decomposes the task into approaching, lifting, and placing stages. Instead of
requiring the executor to infer the entire manipulation procedure from the
global instruction, each subgoal provides a clear intermediate objective.
Consequently, the executor can adapt its behavior according to the current
task stage and maintain consistent progress throughout the execution process,
showing the advantage of hierarchical task decomposition.

Figure~\ref{case}(c) illustrates a more complex multi-object manipulation
scenario, where the robot needs to place two moka pots onto the stove. The
planner first guides the executor to approach and lift the first pot, followed
by placing it on the stove, and then generates new subgoals for manipulating
the second pot. This case highlights that HiRoC can dynamically organize
multiple manipulation steps and handle tasks with repeated object-level
operations. By combining planner-generated subgoals with subgoal-conditioned
executor optimization, HiRoC provides reliable guidance for long-horizon
decision making and achieves robust task completion.
\end{document}